# Margin-Based Feed-Forward Neural Network Classifiers

Han Xiao, Xiaoyan Zhu

*Abstract*—Margin-Based Principle has been proposed for a long time, it has been proved that this principle could reduce the structural risk and improve the performance in both theoretical and practical aspects. Meanwhile, feed-forward neural network is a traditional classifier, which is very hot at present with a deeper architecture. However, the training algorithm of feed-forward neural network is developed and generated from Widrow-Hoff Principle that means to minimize the squared error. In this paper, we propose a new training algorithm for feed-forward neural networks based on Margin-Based Principle, which could effectively promote the accuracy and generalization ability of neural network classifiers with less labelled samples and flexible network. We have conducted experiments on four UCI open datasets and achieved good results as expected. In conclusion, our model could handle more sparse labelled and more high-dimension dataset in a high accuracy while modification from old ANN method to our method is easy and almost free of work.

## I. Introduction

**N**EURAL network, especially feed-forward back propagation neural network (BPNN), has been a classical classifier. Recently deep architecture of neural network is hot for both application and theory. However, the training algorithm of feed-forward neural network, no matter the shadow or the deep, is according to the Widrow-Hoff Principle which means to minimize the squared error with some weight regulation items. This kind of learning algorithms need lots of labelled samples and are tend to be overfitting. The generalization ability of these neural networks is limited by the Widrow-Hoff Principle and least square error optimization procession, so we would like to employ the Margin-Based Principle instead of Widrow-Hoff Principle to obtain a better generalization ability, which is the motivation of our novel model.

Margin-Based Principle is proposed by Vapnik, it could deal with few labelled samples and would become sparse learning structure. The perceptron that could be treated as linear classifier, shares the disadvantages of neural network, since it also needs lots of labelled data and is tend to be overfitting. But when the Max-Margin Principle is applied to perceptron or linear classifier, SVM is born to overcome the disadvantages of Widrow-Hoff Principle, thus SVM could tolerate less labelled data and gain a better generalization ability. We are inspired by above facts, hence we have applied the Margin-Based Principle to the neural network training process and a novel learning model is proposed that can abandon part defects of traditional feed-forward neural networks.

In this paper, we treat feed-forward neural network as a two-step process, the process of input layer to hidden layers could be treated as feature abstraction and the process of hidden layers to output layer could be treated as classification. In the feature abstraction process, feed-forward neural network makes use of linear regression learners to abstract the samples to a new feature space, so we apply Min-Margin Principle to this process to minimize the error of regression or we say the error of abstraction for improving the ability of feature abstraction. While in the classification process, feed-forward neural network makes use of linear classifiers to discriminate, so we apply Max-Margin Principle to this process to minimize the structural risk. Above, we combine the two processes to an optimization problem which is our model.

Our experiments are conducted on four UCI open datasets, which are Banknote Authentication, MAGIC Gamma Telescope, ISOLET and FarmAD. They cover many domains. The results of the experiments prove our novel training algorithm is better than traditional one that ANN, and we can draw a conclusion that Margin-Based Neural Network could achieve better accuracy with less labelled data while modification of program from traditional ANN to our method is almost free of work. Noted that going to deep is also a principle to improve feed-forward neural networks, but it focuses on the refinement of network structures, and we focus on the principle to train the neural network. Our principle could benefit both the shadow and deep model.

As to the application of our model, both the big labelled dataset and the relatively little labelled dataset could totally benefit from our model. Even in the big data time, labelling is also a hard process that costs much. The area such as recommendation system, log analysis in long-tailed retrieval system, medicine and other subjects remote from computer science , also suffer from the problem that little labelled data and huge predicting data. Our method could use less labelled data to achieve better accuracy, for this point, above area could be promoted. Besides, our model could replace ANN method almost without much work in practical programming.

The main contribution of this paper includes:
1) We proposed a novel training principle for feed-forward neural networks, based on Margin-Based Principle.

Xiao is with the State Key Lab. of Intelligent Technology and Systems, National Lab. for Information Science and Technology, Dept. of Computer Science and Technology, Tsinghua University, Beijing 100084, PR China (e-mail:xiaoh12@mails.tsinghua.edu.cn)

Zhu is with the State Key Lab. of Intelligent Technology and Systems, National Lab. for Information Science and Technology, Dept. of Computer Science and Technology, Tsinghua University, Beijing 100084, PR China (e-mail:zxy-dcs@tsinghua.edu.cn)

2) An training algorithm is proposed to solve the optimization problem, and it is a novel training algorithm for feed-forward neural networks.

In section II, we review the related work. In section III, we explore two processes in feed-forward neural network and how Margin-Based Principle could be applied into the two processes. In section IV, we propose our novel method and solution for optimization targets. Experiments are followed and the final section is the conclusion.

## II. RELATED WORK

Neural network is one of early artificial intelligence models, and now becomes a large branch of learning algorithms. What's this paper focus about is the most famous one that feed-forward back prorogation classification networks with non-linear mapping. There are three kinds of most popular ways to promote feed-forward neural network, which are to add weight regularity item to the optimization target, to prune the surplus links and to go into deep architectures. The first way is to refine the weights of the neural network and the other ways are to revise the structure of it. [1] had studied the effect of weight decay and states that it has two effects, suppressing any irrelevant of the weights and improving the generalization ability. [2] had studied the pruning of RBF Neural Network, which firstly introduces the concept of significance of the hidden neurons and then uses it in the learning algorithm to realize parsimonious networks.

Recently such stated in [3], going to deep catches many eyes, since not only just adding the hidden layers could gain an improvement in performance, but deep neural networks can also automatically select features and amazingly complete the comprehension missions. [4] had applied deep network into natural languages, and many works such as [5] had applied deep network into image processing, deep learning is one of the hottest topic in today's AI. Before [6] and [7] proposed the fast unsupervised or supervised methods, multi-layer neural networks are hard to train, this kind of difficulty is analysed in [8] and [9], for the reason that the optimization process is often stuck into the local optima. The work in this paper is a contribution independent from deep learning, or we say, our work is a kind of principle, which could both work for shadow or deep architecture of feed-forward neural networks. As to the difficult of multilayer training algorithms which is applied with our principle, it could also be solved by the same kind of deep learning tricks, introduced by above works. In a word, deep learning is a kind of contribution to the architecture of neural networks and this paper is a kind of contribution to the training principle. Deep architecture and our principle could joint to generate a new powerful model that should be more competent than these two methods alone.

The famous Margin-Based model is Support Vector Machine (SVM) proposed by Vapnik and SVM could have many advantages such as small sample learning and sparse learning structures. However, Margin-Based Principle could be applied to many models to achieve the similar benefits of SVM. [10] had applied the Max-Margin Principle into the Markov Networks, and [11] had applied the Max-Margin

Fig. 1. Two Processes in Two Layers Feed-Forward Neural Networks

Principle into classification of data with absent features. Both of these two works belong to supervised learning. [12] had applied Margin-Based Principle into feature selection and [13] had applied Max-Margin Principle into Clustering. Both of these two works belong to unsupervised learning. Recently, [14] had introduced this principle to on-line learning for Markov Logic Networks, and [15] had introduced it to early event detection.

As above, Margin-Based Principle can also be applied into feed-forward neural networks, which is one of the contributions of this paper. Our work could be applied into neural networks together with weight decay, link pruning and deep architectures. Besides our work can be used in practical usage by directly replacing ANN almost without work.

## III. TWO PROCESSES IN FEED-FORWARD NEURAL NETWORKS

The feed-forward neural network could be treated as two processes,that abstraction process and classification process. Take two-layers feed-forward neural network as example, as Fig. 1 illustrates.

The process of input layer to hidden layers corresponds to abstraction process. In this stage, each hidden neuron is treated as a linear regression learner to fit some parts of the data, and hidden neurons deal the regression results with non-linear function to get its output. In this process, the features of the samples could be converted into new space, where the hidden neurons play a role as the basis. As Fig. 1 illustrates, the hidden neuron $H_1$ catches the characteristic of the black samples, and the hidden neuron $H_2$ catches the characteristic of the white samples, and other hidden neuron like $H_n$ catches some aspect of samples. All of them could abstract the original space and data distribution into an abstraction space where data manifold could be analysed easily.

The process of hidden layers to output layer corresponds to classification process. In this stage, each output neuron is treated as a linear classifier to discriminate different classes of data, and output neurons also deal the classification result with non-linear functions to get its output. As Fig. 1 illustrates, the output neuron $O_1$ is a hyper-plane to separate the abstracted white samples and abstracted black samples.

In deeper architecture, each layer could play a role as abstraction process or classification process. And next, we will discuss about how to train different layers as different roles.

### A. Abstraction Process And Min-Margin Principle

As above stated, the abstraction process makes use of linear regression learners to catch different characteristics of samples. In the traditional way, the process is optimized by least square errors, as followed:

$$\min \quad J = (<\vec{w}, \vec{x}> -y)^2$$

However, Margin-Based Principle minimizes the margin not the distance, so Min-Margin Principle takes the minimum

Fig. 2. Least square error and Min-Margin Principle in abstraction process, (a) shows the least square error and (b) shows the margin-based principle error.

Fig. 3. Least square error and Max-Margin Principle in Classification Process

distance from samples to regression linear plane, the optimized target as followed:

$$\min \quad J = \frac{(<\vec{w}, \vec{x}>)y}{|\vec{w}|}$$

In the equation, the $y$ is the label of the sample for some linear abstraction regression learners, and the labels could be learned by the optimization process.

As Fig. 2 shows, the graph (a) represents the least square error principle and the graph (b) represents the Min-Margin Based Principle. Least square error would be tend to be effected by coordinate geometry, for different coordinate system that may be a rotated one would give out a different regression linear hyper-plane. However, the Min-Margin Principle is geometric invariant and more essential, as one of the reasons why the abstraction process would be promoted.

### B. Classification Process And Max-Margin Principle

As above stated, the classification process should make use of linear classifiers to discriminate different classes, In traditional way, the linear hyper-plane could stop at any suitable position and for this reason, the network would be quickly stuck into overfitting problem. While the Max-Margin Principle could effectively reduce the structural risk and improve the generalization ability, thus Max-Margin Principle takes the idea as followed optimization problem.

$$\max \quad J = \frac{(<\vec{w}, \vec{x}>)y}{|\vec{w}|}$$

As Fig. 3 shows, the solid line corresponds to least square optimization and the dashed line corresponds to Max-Margin Principle optimization. When more testing samples come in, the dashed line could be better than the solid one. Above, Max-Margin Principle could promote the classification process.

### IV. MARGIN-BASED PRINCIPLE FOR FEED-FORWARD NEURAL NETWORKS

The input layer to hidden layers is the abstraction process where we should apply the Min-Margin Principle, and the hidden layers to output layer is the classification process where we should apply Max-Margin Principle. It means that all the hidden layers should be an abstraction process where Min-Margin Principle works and the last layer that corresponds to the output should be a classification process where Max-Margin Principle is applied into.

The feed-forward neural network has $M$ hidden layers each of which has $M_H$ hidden neurons and $N$ output neurons, the number of labelled samples is $T$. So we combine the two principles in one optimization problem, as maximizing followed formulations:

$$Objective = \sum_{t=1}^{T}\{\sum_{i=1}^{N} \frac{<\vec{w}_i^{out}, \vec{y}^{M+1}> t_i}{|\vec{w}_i^{out}|} - \lambda \sum_{m=1}^{M}\sum_{j=1}^{M_H} \frac{<\vec{w}_j^m, \vec{y}^m> y_j^{m+1}}{|\vec{w}_j^m|}\} \quad (1)$$

In above formula, the $\vec{w}_i^{out}$ is the weight vector from last hidden layer to the $i$-th neuron in output layer, and the $\vec{w}_i^m$ is the weight vector from $(m-1)$-th hidden layer to the $i$-th neuron in $m$-th hidden layer. $t_i$ is the output of $i$-th neuron in output layer, and $y_j^m$ is the output of $j$-th neuron in $m$-th hidden layer. $\vec{x}_t$ is the input vector, and $\vec{y}^m$ is the output vector of $(m-1)$-th hidden layer. $\vec{y}^m$ is composed by $y_j^m$. $\lambda$ is an hyper-parameter in the training algorithm. We noted the above formula as $J$.

The first item in above formula corresponds to the Max-Margin Principle for classification process, and the second item corresponds to the Min-Margin Principle for abstraction process.

What we modified is the training algorithm comparing to ANN, and the inference algorithm is as same as traditional feed-forward neural network.

$$y_j^{m+1} = \sigma(<\vec{w}_j^m, \vec{y}^m>)$$

And for brief notation, we have $\vec{y}^1 = \vec{x}$, $\vec{y}^{M+2} = \vec{t}$ and $\vec{w}^{out} = \vec{w}^{M+1}$. $y^m$ means the vector of outputs of $(m-1)$-hidden layer, it is composed by $y_j^m$.

However, the output of non-linear function must be a symmetric expression for both maximum positive value and minimum negative value, so we define it and its derivative as below:

$$\sigma(x) = \frac{1}{1+\exp(-x)} - 0.5 \quad (2)$$
$$\sigma'(x) = (0.5 + \sigma(x))(0.5 - \sigma(x)) \quad (3)$$

The training algorithm adopts gradient ascent for the optimization target as followed:

$$J_t = \sum_{i=1}^{N} \frac{<\vec{w}_i^{out}, \vec{y}^{M+1}> t_i}{|\vec{w}_i^{out}|} - \lambda \sum_{m=1}^{M}\sum_{j=1}^{M_H} \frac{<\vec{w}_j^m, \vec{y}^m> y_j^{m+1}}{|\vec{w}_j^m|}$$

Gradient ascent algorithm is used, so we should work out the partial derivative of the target. In order to express these formula in a brief way, firstly we introduce a $\delta$ function and it can be worked out iteratively.

$$\delta_{j,M}^{out} = \sum_{s=1}^{N} \frac{t_s}{|\vec{w}_s^{out}|} \vec{w}_s^{out}(j) \sigma'(<\vec{w}_j^M, \vec{y}^M>) \quad (4)$$

$$\delta_{j,m-1}^{out} = \sum_{s=1}^{M_H} \delta_{s,m}^{out} \vec{w}_s^m(j) \sigma'(<\vec{w}^{m-1}, \vec{y}^{m-1}>) \quad (5)$$

$$\delta_{j,m}^m = \sum_{s=1}^{N} \frac{y_s^{m+2}}{|\vec{w}_s^{m+1}|} \vec{w}_s^{m+1}(j) \sigma'(<\vec{w}_j^m, \vec{y}^m>) \quad (6)$$

$$\delta_{j,m-1}^m = \sum_{s=1}^{M_H} \delta_{s,m}^m \vec{w}_s^m(j) \sigma'(<\vec{w}^{m-1}, \vec{y}^{m-1}>) \quad (7)$$

Then, we define a $\gamma$ function.

$$\gamma_j^m = \frac{y_j^{m+1}}{|\vec{w}_j^m|}\vec{y}^m - \frac{<\vec{w}_j^m, \vec{y}^m> y_j^{m+1}}{|\vec{w}_j^m|^3}\vec{w}_j^m$$
$$+ \frac{<\vec{w}_j^m, \vec{y}^m> \sigma'(<\vec{w}_j^m, \vec{y}^m>)}{|\vec{w}_j^m|}\vec{y}^m \quad (8)$$

With the form of $\delta$ function, we could reduce the computational complexity and express the gradient, very briefly.

$$\frac{\partial J_t}{\partial \vec{w}_i^{out}} = \frac{t_i}{|\vec{w}_i^{out}|}\vec{y}^{M+1} - \frac{t_i <\vec{w}_i^{out}, \vec{y}_{M+1}>}{|\vec{w}_i^{out}|^3}\vec{w}_i^{out} \quad (9)$$

$$\frac{\partial J_t}{\partial \vec{w}_i^m} = \delta_{i,m}^{out}\vec{y}^m - \lambda \sum_{s=m+1}^{M} \delta_{i,m}^s \vec{y}^m - \lambda \gamma_j^m \quad (10)$$

In above formula, $\sigma'$ is the derivative of the non-linear function. With the derivative of the target, we can obtain the updating equation.

$$\vec{w}_i^{out} = \vec{w}_i^{out} + \alpha * \frac{\partial J_t}{\partial \vec{w}_i^{out}} \quad (11)$$

$$\vec{w}_j^m = \vec{w}_j^m + \alpha * \frac{\partial J_t}{\partial \vec{w}_j^m} \quad (12)$$

$\alpha$ is the learning rate. So, the training algorithm is achieved. This method in some special case may be stuck into local optima. This is the common problem for neural networks.

The temporal complexity for this algorithm to train one sample is $O(|NodeNumber|^2|LayerNumber|^2)$. For the reason that the $|LayerNumber|$ is fixed and small, our algorithm is as fast as traditional ones.

Traditional feed-forward neural network would have a problem that is the saturation of this sigmoid function, and for this reason, the training is hardly possible in high-dimension dataset. However, in this optimization of our model, this problem is solved, by our part gradient items that do not involve the derivative of sigmoid function. So our model could solve high-dimension problems directly and naturally while traditional one must use other technical. This is one of advantages of our model.

## V. EXPERIMENTS

We have conducted two groups of experiments, one group for the effectiveness of Margin-Based Feed-Forward Neural Network, another group for the studies of the network structure or we say hyper-parameters.

### A. Datasets

This paper has selected four UCI open datasets as experiment datasets. They are Banknote Authentication, MAGIC Gamma Telescope, ISOLET and FarmAD. The rule we choose dataset is to verify our methods in different settings and different domains.

In order to make training dataset and testing dataset, we split the dataset randomly, and the result of performance is the average of at least five times split testing. So our data about performance is robust and believable in the statistic view.

TABLE I.
Accuracy of Four Models

| Dataset | Percent | ANN | SVM | AdaBoost | Ours |
|---|---|---|---|---|---|
| BankNote | 10 % | 0.99 | 0.98 | 0.95 | **0.99** |
| BankNote | 2 % | 0.95 | 0.96 | 0.94 | **0.97** |
| BankNote | 1 % | 0.91 | 0.91 | 0.81 | **0.95** |
| Magic | 1% | 0.70 | 0.76 | **0.77** | 0.73 |
| Magic | 0.5% | 0.68 | 0.72 | 0.72 | **0.73** |
| Magic | 0.25% | 0.67 | 0.67 | 0.70 | **0.73** |
| ISOLET | 10 % | 0.89 | 0.90 | 0.85 | **0.91** |
| ISOLET | 3.33 % | 0.84 | 0.81 | 0.81 | **0.87** |
| ISOLET | 2 % | 0.82 | 0.81 | 0.82 | **0.85** |
| FarmAD | 20 % | 0.54 | 0.83 | 0.78 | **0.87** |
| FarmAD | 10 % | 0.53 | 0.81 | 0.78 | **0.84** |
| FarmAD | 3.33% | 0.51 | 0.76 | 0.69 | **0.80** |

**Banknote Authentication.** The data is extracted from images that were taken from genuine and forged banknote-like specimens and Wavelet Transform tool were used to extract features from images. There are 1,372 items and 5 attributes for binary classes.

**MAGIC Gamma Telescope,** The data is generated to simulate registration of high energy gamma particles in a ground-based atmospheric Cherenkov gamma telescope using the imaging technique. There are 19,020 items and 11 attributes for binary classes.

**ISOLET.** The dataset is a real speech recognition dataset, in which we must distinguish between vowels and consonants. there are 7,797 samples with 618 attributes.

**FarmAD,** It is a real text classification dataset, with 3,751 documents and each has 4,200 word bag attributes. This data was collected from text ads found on twelve websites that deal with various farm animal related topics. Information from the ad creative and the ad landing page is included. The labels are based on whether or not the content owner approves of the Ad.

### B. Evaluation Of Effectiveness

For evaluation of effectiveness of our training algorithms, we choose three baselines, and the baselines and our model are listed as below.
1) Two-Layers Neural Network with the same suitable structure as our models, notated as ANN. This model is used to prove our effectiveness of training principle. It's implemented by ourselves.
2) Support Vector Machine (SVM) with Radio Basis Function Kernels. It is implemented by Weka using SMO algorithm, notated as SVM-RBF. This model is used to prove our effectiveness of Min-Margin Principle for Abstraction Process.
3) AdaBoost, which is implemented by Weka, notated as AdaBoost. This model is used

Fig. 4. Accuracy for different hidden nodes in 3.33% sparse ISOLET dataset. The x-axis means hidden node number, the y-axis means accuracy. The blue line corresponds to ANN while the green line corresponds to Ours.

- to prove our effectiveness of Max-Margin Principle for Classification Process.
4) Our Model, with the same suitable network structure with the first baseline, but using different training algorithms while the inference algorithm is same, notated as Ours.

What we want to evaluate is how the effectiveness of all the four classifiers in the different scalable training datasets. For example, take 5% as labelled training data of the dataset and others as testing data, we can evaluate the performance of these four classifiers, and take 10% as labelled training data and others as testing, we can also evaluate the performance of these four classifiers. However, always in practical usage, labelled data is so few while the data to be predicted is so huge, hence we focus to evaluate the performance is in the situation of suitable labelled training data.

Table 1 illustrates all the performance of this experiment in four open UCI datasets. The 'Labelled Percent' means the percent of dataset for training and others for testing. From this results, we can conclude some key points:

1) Compared with ANN, it's proved that our training algorithm is better than traditional model. The reason is that we apply the Margin-Based Principle into Feed-Forward neural network and Margin-Based Principle works. Our method is effective.
2) Compared with SVM-RBF, we can see our model is more effective, because the mapping of kernel methods doesn't consider the data manifold, and our abstract process which is promoted by Min-Margin Principle that considers the data manifold.
3) Compared with AdaBoost, we can see our classification process which is promoted by Max-Margin Principle works well, because the AdaBoost also uses many classifiers to consider the data manifold in abstraction process but we are different in classification process. Our classification with Max-Margin Principle works well.

In conclusion, our model works obviously well, especially for small training and big testing datasets. This matches our theory about our model, the result is as good as expected.

### C. Evaluation Of Network Structure

In order to study about effect of the network structure, we conduct experiments on the dataset ISOLET with a suitable sparse training dataset sparsity of which is 3.33%. We construct the same network structure but with different training algorithm to learn. The result is showed in Fig. 4. In the experiment, different neural networks have different hidden neuron number, meaning denser network has, more hidden neuron number is,

In the Fig. 4, more x-axis is, denser the network is, and the y-axis is the accuracy.

For the results in Fig. 4, we can see some key points:

1) There exists a suitable hidden node number, less than which leads to unfitting problem while more than which leads to overfitting problem.
2) In an overview of trends and values, our method outperforms ANN and also share almost same trend with ANN, which means the hyper-parameter of ANN could be transducted to our method. In practical usage, our method could replace ANN almost without modifying programs while the performance would be improved. The reason for this point should lie that the inference process and network structure is the same between our model and ANN.

## VI. CONCLUSION

In this paper, we formulate Margin-Based Principle into the training algorithms of Feed-Forward Neural Networks, and two kinds of processes is studied in detailed. In abstraction process Min-Margin Principle is applied and In classification process Max-Margin Principle is applied. Based on this point, we propose the training algorithm to solve the Margin-Based optimization problem. All the theory is evaluated in real open datasets, and good results are achieved as expected. In conclusion, our model could handle more sparse labelled datasets and more high-dimension datasets in a high accuracy while modification from old ANN method to our method is easy and almost free of work.